\documentclass[numsec,webpdf,modern,medium]{oup-authoring-template}

\graphicspath{{Fig/}}

\theoremstyle{thmstyleone}%
\theoremstyle{thmstyletwo}%
\theoremstyle{thmstylethree}%

\usepackage{seqsplit} 
\usepackage{xurl} 
\usepackage{endnotes}
\begin{document}

\journaltitle{Journal Title Here}
\DOI{DOI HERE}
\copyrightyear{2025}
\pubyear{2025}
\access{Advance Access Publication Date: Day Month Year}
\appnotes{Paper}

\firstpage{1}

\title[Quantitative Intertextuality]{Quantitative Intertextuality from the Digital Humanities Perspective: A Survey}

\author[1,$\ast$]{Siyu Duan}

\authormark{Siyu Duan}

\address[1]{\orgdiv{School of Foreign Languages}, \orgname{Shanghai Jiao Tong University}, \orgaddress{\street{800 Dongchuan Road}, Minhang District, \postcode{200240}, \state{Shanghai}, \country{China}}}

\corresp[$\ast$]{Corresponding author. \href{email:email-id.com}{duansiyu@pku.edu.cn}}

\received{Date}{0}{Year}
\revised{Date}{0}{Year}
\accepted{Date}{0}{Year}



\abstract{The connection between texts is referred to as intertextuality in literary theory, which served as an important theoretical basis in many digital humanities studies. Over the past decade, advancements in natural language processing have ushered intertextuality studies into the quantitative age. Large-scale intertextuality research based on cutting-edge methods has continuously emerged. This paper provides a roadmap for quantitative intertextuality studies, summarizing their data, methods, and applications. Drawing on data from multiple languages and topics, this survey reviews methods from statistics to deep learning. It also summarizes their applications in humanities and social sciences research and the associated platform tools. Driven by advances in computer technology, more precise, diverse, and large-scale intertext studies can be anticipated. Intertextuality holds promise for broader application in interdisciplinary research bridging AI and the humanities.}



\maketitle

\section{Introduction}
Intertextuality refers to the connection between texts, a concept introduced by French literary critic Julia Kristeva \citep{kristeva1980word}. It plays an important role in literary studies and textual analysis. Intertextuality can occur across different textual levels, such as words, sentences, and documents, and take various forms, including semantic connections, rhythmic similarities, or shared themes and narrative structures. Initially confined to textual connections, the concept has expanded to include relationships across other modalities, such as sound and image. Nevertheless, text-based intertextuality research remains dominant.

Before the formal introduction of intertextuality, studies of textual connections were already underway in literary scholarship. As early as 1818, Chinese scholars attempted to list textual parallels between \textit{Han Shi Wai Zhuan} and several major works from the Axial Age \citep{shike}. Such manual efforts continued into the 21st century \citep{Hezh2024}. Western literary studies followed a similar path, with extensive work on allusions and influences in Latin literature. For instance, the enduring analysis of how Virgil's \textit{Aeneid} influenced Lucan's \textit{Pharsalia}. Despite considerable humanistic efforts to exhaustively trace intertextual links, only a small fraction of possible connections were uncovered. 

\begin{figure*}
    \centering
    \includegraphics[width=1\linewidth]{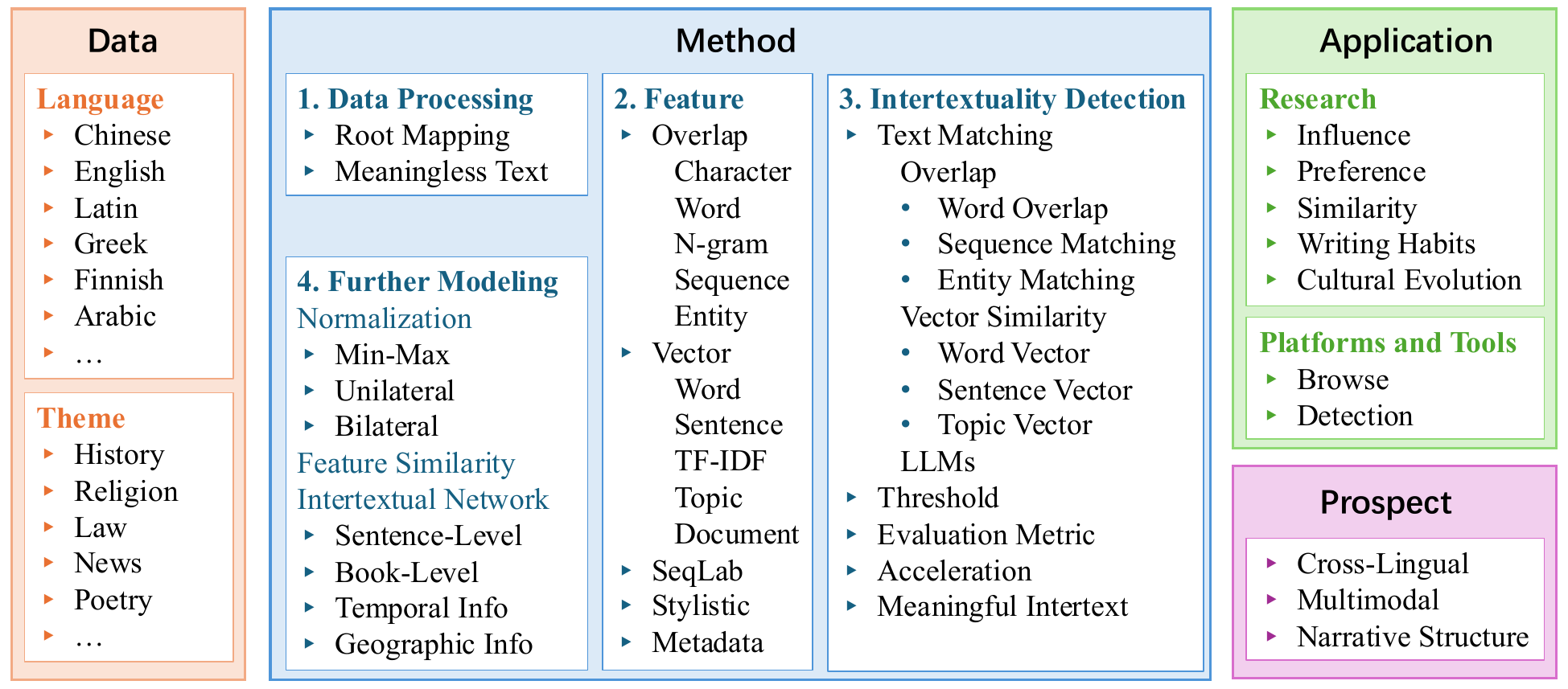}
    \caption{Technical framework of quantitative intertextuality}
    \label{fig:framework}
\end{figure*}

The digital methods have brought unprecedented advances to intertextuality research. With the advance of natural language processing, quantitative intertextuality studies have expanded in both breadth and depth. Computers can now identify millions of textual parallels within hours. In 2018, one study reported around 200,000 intertextual pairs (66,000 sub-clusters, 202,000 texts) \citep{sturgeon2018unsupervised}. Using deep learning models and large-scale vector retrieval engines, this number grew to 2.6 million pairs in 2023 \citep{duan2023disentangling} and 14 million by 2024 \citep{wang2024evol}. Methodologically, early studies (pre-2020) largely relied on character overlap to detect observable textual similarities. More recent approaches employ deep neural networks to match texts in high-dimensional vector spaces, capturing more abstract intertextual relationships. These modeling results provide quantitative evidence for humanities and social sciences studies. Today, digital platforms for intertext browsing and detection, which are often enhanced by information retrieval and visualization tools, enable deeper reader engagement.

A book named \textit{Quantitative Intertextuality} \citep{forstall2019quantitative} offers a systematic overview of this field. However, its focus is primarily on Western literature, and it was published before the widespread adoption of deep learning—thus limited in both scope and technical currency. This paper surveys quantitative intertextuality research across multiple languages, including Chinese, English, and Latin, and spans disciplines such as history, religion, literature, law, and journalism. The technical framework for quantitative intertextuality research is summarized in the Fig. \ref{fig:framework}. This paper summarizes the data, methods, and applications of these studies, highlights their commonalities and distinctive features, and discusses challenges and future directions for intertextuality research.

\section{Data}

\begin{table*}
    \renewcommand \arraystretch {1.2}
    \centering
    \caption{Literature categorized by language}
    \begin{tabular}{l|l}
    \hline
    Language     & Literature\\
    \hline
    Chinese & \textit{Classical Chinese}: \\
    &\cite{Xiao2010guji}, \cite{sturgeon2018digital}, \cite{sturgeon2018unsupervised}, \cite{sturgeon2019}, \\
    &\cite{nehrdich2020method}, \cite{Liang2021guji},  \cite{zhou2021p}, \cite{huang2021yinshu},\\
    & \cite{deng-etal-2022-shu}, \cite{li-etal-2022-zhen}, \cite{cheng2023}, \cite{duan2023disentangling}, \\
    &\cite{wang2024evol}, \cite{mcmanus2025measuring}, \cite{fu2025ancienttrd} \\
    &\textit{Classical and Modern Chinese}: \cite{yu-etal-2022-ji} \\
    &\textit{Modern Chinese}: \cite{bothwell2023introducing} \\
    \hline
    English &\textit{Modern English}: \\
    &\cite{smith2013infectious}, \cite{smith2014detecting}, \cite{buchler2014towards}, \cite{wilkerson2015tracing}, \\
    &\cite{smith2015computational}, \cite{cordell2015reprinting}, \cite{williams2015method}, \cite{romanello2016exploring}, \cite{burgess2016legislative}, \\
    &\cite{duhaime2016textual}, \cite{cordell2017fugitive}, \cite{hartberg2017sacred}, \\
    &\cite{chaturvedi2018have}, \cite{linder2020text}, \cite{o2021text}, \cite{shang2021improving}, \\
    &\cite{hatzel2024tell}, \cite{hatzel2024story}, \cite{karjus2025machine}\\
    &\textit{Old English}: \cite{neidorf2019large}\\
    \hline
    Latin & \cite{forstall2011evidence}, \cite{coffee2012intertextuality}, \cite{coffee2013tesserae}, \cite{scheirer2014thesense}, \\
    &\cite{forstall2014modeling}, \cite{bernstein2015comparative}, \cite{moritz-etal-2016-non}, \cite{dexter2017quantitative},\\
    &\cite{gawley2017comparing}, \cite{manjavacas2019feasibility},  \cite{burns2021profiling}, \cite{bothwell2023introducing}\\
    \hline
    Greek&  \cite{lee-2007-computational}, \cite{buchler2013measure},  \cite{moritz-etal-2016-non}, \cite{umphrey2024investigating}\\
\hline
Finnish& \cite{vesanto2017applying}, \cite{vesanto2017system},  \cite{salmi2020reuse}, \cite{janicki2023exploring}\\
    \hline
    Other&  \cite{mohammad2017paraphrase} (Arabic), \cite{ganascia2014automatic} (French), \\
    &\cite{shmidman2018identification} (Hebrew/Aramaic), \cite{karjus2025machine} (Russian) \\
    \hline
    \end{tabular}
    \label{tab:datalang}
\end{table*}

Quantitative intertextuality has been applied to literature in various languages, as categorized by language in Table \ref{tab:datalang}. It is evident that quantitative intertextuality is popular in the study of ancient literature. Compared to modern literature, the distinctive features and inherent limitations of ancient texts make them especially suitable for intertextual research:
\begin{itemize}
\item Lack of writing conventions. Modern literary works adhere to established standards such as quotation marks and reference lists, which are often absent in ancient literature.
\item Missing metadata. For some ancient texts, author identities, places of origin, and dates of creation are unknown and difficult to verify.
\item Missing media information. Modern publications often come with contextual information provided by media channels, such as location, publisher, online attention, and sharing data. In contrast, the dissemination pathways of ancient documents recorded on materials like paper, bamboo, or stone are challenging to reconstruct.
\end{itemize}
Due to these limitations, the analysis of ancient literature often depends primarily on the text itself. Since intertextuality is a universal feature across all texts, it offers a compatible framework for studying ancient works.

Previous studies have explored intertextual connections around various themes, as summarized in Table \ref{tab:datatype}. Some studies utilize multiple types of source materials, which are listed separately under each relevant category. Ancient Roman epic, for instance, possesses both historical and poetic qualities; this paper classifies it under the ``history'' theme. For corpora that incorporate five or more thematic types, they are grouped under the ``comprehensive'' category.

\begin{table*}
    \renewcommand \arraystretch {1.2}
    \centering
    \caption{Literature categorized by theme}
    \begin{tabular}{l|l}
    \hline
    Theme     & Literature\\
    \hline
    History& \cite{coffee2012intertextuality}, \cite{buchler2013measure}, \cite{coffee2013tesserae},  \cite{scheirer2014thesense}, \\
    &\cite{forstall2014modeling}, \cite{bernstein2015comparative}, \cite{dexter2017quantitative}, \cite{neidorf2019large}, \\
    &\cite{burns2021profiling}, \cite{deng-etal-2022-shu}, \cite{yu-etal-2022-ji} \\
    \hline
    Religion&  \cite{lee-2007-computational}, \cite{buchler2014towards}, \cite{moritz-etal-2016-non}, \cite{hartberg2017sacred}, \\
    &\cite{manjavacas2019feasibility}, \cite{nehrdich2020method}, \cite{bothwell2023introducing},\\ &\cite{mcmanus2025measuring}, \cite{umphrey2024investigating}\\
    \hline
    Law& \cite{smith2014detecting}, \cite{wilkerson2015tracing}, \cite{burgess2016legislative}, \\
    &\cite{linder2020text}, \cite{gava2021legislating}\\
    \hline
    News& \cite{smith2013infectious}, \cite{smith2014detecting}, \cite{smith2015computational}, \cite{cordell2015reprinting}, \\
    &\cite{cordell2017fugitive}, \cite{vesanto2017applying}, \cite{vesanto2017system}, \\
    &\cite{mohammad2017paraphrase}, \cite{salmi2020reuse}\\
    \hline
    Poetry& \cite{forstall2011evidence}, \cite{williams2015method}, \cite{shang2021improving}, \\
    &\cite{janicki2023exploring}, \cite{cheng2023}\\
    \hline
    Annotation& \cite{Xiao2010guji}, \cite{huang2021yinshu}, \\
    &\cite{Liang2021guji}, \cite{zhou2021p} \\
    \hline
    Summary& \cite{chaturvedi2018have}, \cite{hatzel2024tell}, \cite{hatzel2024story}\\ 
    \hline
    Comprehensive& \cite{buchler2013measure}, \cite{gawley2017comparing}, \cite{sturgeon2018unsupervised}, \\
    &\cite{shmidman2018identification}, \cite{sturgeon2019}, \cite{o2021text}, \cite{li-etal-2022-zhen}, \\
    &\cite{duan2023disentangling}, \cite{wang2024evol}, \cite{fu2025ancienttrd} \\
    \hline
    Other& \cite{bothwell2023introducing} (Composition), \cite{romanello2016exploring} (Catalogue), \\
    &\cite{dexter2017quantitative} (Drama), \cite{sturgeon2018digital} (Philosophy), \\
    &\cite{duhaime2016textual}, \cite{ganascia2014automatic} (Fiction), \cite{karjus2025machine} (Blog) \\
    \hline
    \end{tabular}
    \label{tab:datatype}
\end{table*}

It can be observed that texts suitable for quantitative intertextuality research share a common characteristic: different versions address the same content. For example, multiple newspapers may cover the same news story; different countries or regions may enact similar laws; various historians may chronicle the same historical period; different translators may render the same religious text; and multiple scholars may provide distinct annotations for the same book. When texts exhibit such characteristics, researchers can prioritize the use of intertextuality modeling analysis.


\section{Calculation Method}
This section outlines the computational methodology for quantitative intertextuality research. It addresses the following key questions: 1) How should textual data be preprocessed to facilitate intertextual modeling? 2) How can appropriate intertextual features be selected from the data? 3) How can parallel intertextuality detection be performed efficiently and accurately? 4) How can similarity be measured based on intertextual features? 5) How can an intertextual network be constructed? This section presents the complete computational pipeline, thereby clarifying the technical framework of this research paradigm.

\subsection{Data Processing}
The data processing pipeline in quantitative intertextuality research should be tailored according to the specific research questions and linguistic features involved. Beyond standard data cleaning and preprocessing, two critical steps for ensuring the accuracy of subsequent intertextuality calculations are root mapping and the handling of meaningless text.

\subsubsection{Root Mapping}
Due to factors such as historical period, geographic region, and inherent language properties, the same semantic unit can appear in different surface forms. For example, in English, a single word may have variations in capitalization, tense, number (singular/plural), or composition (e.g., hyphenation, affixes). In Chinese, a character might have multiple variant forms, and there is a complex mapping relationship between simplified and traditional characters. In intertextuality research, it is often necessary to normalize these variants back to a common root word or character. For digitally well-resourced languages like English and Chinese, open-source external toolkits can be employed to perform root mapping. For low-resource languages, acquiring or creating relevant linguistic resources is necessary.


\subsubsection{Meaningless Text}
When the focus is on ideological connections, such as tracing the dissemination of ideas through texts, researchers prioritize intertextual links involving meaningful content. Therefore, meaningless words often receive special treatment, such as being deleted entirely or having texts with an excessive proportion of them filtered out. These words can be identified using existing external resources, like publicly available stop-word lists, or through methods including part-of-speech tagging and frequency analysis.

\subsection{Intertextual Feature}

Intertextuality is a broad and multifaceted concept. It operates across multiple dimensions, such as semantics, literary style, and metadata, and manifests at various levels, including words, sentences, paragraphs, and documents. Researchers have studied intertextual features from diverse perspectives, and the table \ref{tab:feature} categorizes literature by feature type.

\begin{table*}
    \centering
    \caption{Feature selection in intertextual computation}
    \renewcommand \arraystretch {1.2}
    \begin{tabular}{ll|l}
\hline
\multicolumn{2}{c|}{Feature}&Literature\\
\hline
&Character&\cite{vesanto2017applying}, \cite{vesanto2017system},  \cite{shmidman2018identification}, \\
&&\cite{salmi2020reuse}, \cite{janicki2023exploring}\\
\cline{2-3}
&Word&\cite{Xiao2010guji}, \cite{coffee2012intertextuality}, \cite{coffee2013tesserae}, \\
&&\cite{forstall2014modeling}, \cite{williams2015method}, \cite{moritz-etal-2016-non}, \\
&&\cite{mohammad2017paraphrase}, \cite{chaturvedi2018have}, \cite{manjavacas2019feasibility} \\
\cline{2-3}
Overlap&N-gram&\cite{smith2013infectious},  \cite{smith2014detecting}, \cite{buchler2014towards}, \cite{wilkerson2015tracing}, \\
&&\cite{williams2015method}, \cite{burgess2016legislative}, \cite{duhaime2016textual}, \cite{mohammad2017paraphrase}, \\
&&\cite{sturgeon2018digital}, \cite{sturgeon2018unsupervised}, \cite{sturgeon2019}, \cite{linder2020text}, \\
&&\cite{o2021text},  \cite{li-etal-2022-zhen}, \cite{mcmanus2025measuring} \\
\cline{2-3}
&Sequence&\cite{Xiao2010guji}, \cite{ganascia2014automatic}\\
\cline{2-3}
&Entity&\cite{chaturvedi2018have}, \cite{mohammad2017paraphrase}, \cite{deng-etal-2022-shu} \\
\hline
&Word& \cite{manjavacas2019feasibility}, \cite{nehrdich2020method}, \cite{burns2021profiling}, \cite{li-etal-2022-zhen}\\
\cline{2-3}
&Sentence&\cite{duan2023disentangling}, \cite{wang2024evol}, \cite{Liang2021guji}, \cite{li-etal-2022-zhen}, \\
&&\cite{cheng2023}, \cite{yu-etal-2022-ji}, \cite{fu2025ancienttrd} \\
\cline{2-3}
Vector&TF-IDF&\cite{lee-2007-computational}, \cite{sturgeon2018digital}, \cite{manjavacas2019feasibility}, \\
&&\cite{deng-etal-2022-shu}, \cite{shang2021improving} \\
\cline{2-3}
&Topic&\cite{scheirer2014thesense}, \cite{mohammad2017paraphrase} \\
\cline{2-3}
&Document&\cite{hatzel2024story}, \cite{hatzel2024tell}, \\
&&\cite{umphrey2024investigating}, \cite{karjus2025machine}\\
\hline
&SeqLab& \cite{zhou2021p}, \cite{bothwell2023introducing}\\
\cline{2-3}
Other&Stylistic&\cite{neidorf2019large}, \cite{forstall2011evidence}, \cite{dexter2017quantitative}\\
\cline{2-3}
&Metadata&\cite{huang2021yinshu},  \cite{romanello2016exploring}\\
\hline

    \end{tabular}
    \label{tab:feature}
\end{table*}

\subsubsection{Overlap Feature}
Overlap features refer to directly observable identical elements between two texts, such as matching characters, words, n-grams, sequences, and entities. These features are visually identifiable and highly interpretable.

Here, “characters” refer to those in alphabetic writing systems. The smallest semantic unit is the word, which is composed of characters; characters themselves carry no inherent meaning. When text data quality is low, for instance, too many OCR errors \citep{vesanto2017applying, salmi2020reuse, vesanto2017system}, unordered character-based features are more suitable. Using ordered character sequences under such conditions may result in poor matching performance due to frequent character-level errors. 
Many languages exhibit orthographic variations, such as character variants in Chinese or differences in capitalization and number in English. For well-resourced languages, external toolkits can normalize such variations by mapping words to their root forms. However, for languages with limited resources, such as Finnish \citep{janicki2023exploring} and Hebrew-Aramaic \citep{shmidman2018identification}, which also exhibit significant spelling variations, overlap measures at the character level are more appropriate.

Word overlap, n-gram overlap, and sequence matching are highly generalizable methods that have been successfully applied across many languages. The minimal unit for these features is a word in alphabetic systems or a character in logographic systems. 
While word overlap ignores word order, n-gram matching introduces limited order constraints. An n-gram is a short consecutive sequence of words (or characters in logographic systems), where n typically ranges from 2 to 5 and seldom exceeds 10. A common strategy is to first detect text pairs sharing the same n-grams (often accelerated via hashing algorithms) and then apply sequence scoring rules for finer filtering. This two-step process balances efficiency and accuracy. Sequence features, in contrast, are derived directly from full-sequence alignment. 

Texts rich in named entities, such as historical documents \citep{deng-etal-2022-shu} or narratives \citep{chaturvedi2018have}, are well-suited for entity overlap measures, as they contain many proper nouns that are not adequately handled by standard semantic matching approaches.

\subsubsection{Vector Feature}
Vector features convert discrete characters into continuous vectors, mapping semantics into a high-dimensional space to capture more abstract semantic representations. Commonly used vector types include word vectors, sentence vectors, and paragraph- or document-level vectors.

Word vectors generally cannot be used in isolation; they require additional rule design to measure text similarity at the n-gram level and above \citep{burns2021profiling, li-etal-2022-zhen}. Sentence vectors encapsulate more complex semantics and can help identify richer, more abstract intertextual connections.
Both word and sentence vectors can be obtained by using pre-trained language models. Alternatively, these models can be further fine-tuned on a specific target corpus to produce vector representations that are better suited to particular research needs.

TF-IDF and topic vectors are appropriate for matching longer texts, such as paragraph or document. Topic vectors can be generated using models like LDA or LSI, which capture implicit connections through thematic features \citep{scheirer2014thesense}. 
Document-level vector features are suitable for capturing linkages in narrative structures \citep{hatzel2024story, huang2021yinshu}, or for processing entire texts through a generative LLM \citep{umphrey2024investigating}.

The disadvantage of vector-based features is their poor interpretability. The semantics are calculated in a latent space, a characteristic that may be unsatisfactory to traditional scholars. Nevertheless, it is undeniable that vector methods have significantly expanded the depth and breadth of intertextuality research, enabling the detection of more complex and diverse intertextual relationships.

\subsubsection{Stylistic Feature}
Stylistics emphasizes the form of text rather than the meaning. 
Related studies \citep{dexter2017quantitative, neidorf2019large, forstall2011evidence} have employed stylistic features including textual cadence, sentence structure, entity distribution, and patterns of function word usage, etc. When different texts display intentional or unintentional resemblances in formal aspects, a subtle relationship emerges between them. In genres with strict rhythmic demands, like poetry and drama, stylistic features are particularly useful.

Using stylistic features for intertextual research offers two main advantages: non-sparsity and diversity. Semantically similar parallel texts are relatively scarce in literature: two documents may share only a few parallel intertexts, which is often insufficient for rigorous quantitative analysis. In contrast, stylistic features are inherent to the text and exhibit a denser distribution. Moreover, they can be tailored to specific languages and genres, yielding a wide variety of measurable traits. For a given research question, the convergence of multiple stylistic indicators strengthens the credibility of the findings.

A limitation of stylistic features is their strong dependence on language type. It is often challenging to transfer stylistic criteria from one language to another or to draw inspiration from studies in non-target languages. Consequently, the design of stylistic features must be dynamically adapted according to variations in language type and genre.

\subsubsection{Metadata}
Metadata refers to data that describes other data, offering contextual details such as format, author, creation time, and location. Modern scholarly texts follow standardized citation formats, and citation analysis builds upon these established citation networks. For example, \cite{citron2015patterns} detected text reuse in contemporary academic literature and performed further analysis using metadata such as author, references, topic, and national affiliation of the paper. 

In contrast, ancient literature lacks a standardized citation format, which requires citation identification. Citations appear in two forms: one involves restating or paraphrasing the original content, while the other merely mentions the name of an author or a document without quoting the content directly. The former can be identified through semantic similarity, whereas the latter requires tailored approaches.

For the latter type of citation, Named Entity Recognition (NER) is useful. For instance, \cite{huang2021yinshu} extracted citation relationships from annotated ancient Chinese books by recognizing book titles via NER. Similarly, \cite{romanello2016exploring} identified citation relationships in catalog books using author names and citation scopes (such as volume and page numbers). Since entities like personal names and book titles often have multiple aliases, entity disambiguation should be performed after NER to achieve more accurate citation analysis.

\subsubsection{Sequence Labeling}
Sequence tagging is not used for matching intertexts, but for identifying the presence of certain types of intertexts in text. For example, identifying the location of quoted text in a text \citep{zhou2021p}, or identifying rhetorical texts in a parallel pattern \citep{bothwell2023introducing}.

\subsection{Parallel Intertextuality Detection}

Listing texts with intertextual connections is an important component of traditional intertextuality research. Originally reliant on extensive manual effort, this process has undergone transformative advances within computer technology. Researchers have explored many computational approaches for detecting parallel intertexts, evolving from rule-based matching to deep learning models, significantly improving the accuracy and diversity of detection. This section reviews existing methodologies from five key perspectives: text matching, threshold determination, evaluation metrics, computational acceleration, and the identification of meaningful intertexts.

\subsubsection{Text Matching}
This paper categorizes intertext recognition techniques into two main types: those based on lexical overlap and those based on vector similarity. Overlap-based methods can be further divided according to whether they account for word order: these include word overlap and sequence matching. In addition, entity matching can be integrated with both types to enhance performance.

\textbf{Word Overlap}.
Methods based on word overlap ignore word order and typically identify parallel text pairs that contain at least n identical words or shared n-grams. For instance, some studies directly retrieve texts that share two or more identical words \citep{coffee2012intertextuality, coffee2013tesserae, forstall2014modeling}. While straightforward, this approach often requires secondary filtering to eliminate unqualified matches.

\textbf{Sequence Matching}.
Sequence matching methods take word order into account. Researchers typically define sequence similarity based on data characteristics and research objectives. For example, \citet{sturgeon2018unsupervised} first identified all strings sharing identical 4-grams, then examined their contexts to refine match boundaries and filter low-quality alignments. \citet{li-etal-2022-zhen} designed scoring functions grounded in n-gram matching to quantify sequence similarity. \cite{wilkerson2015tracing, linder2020text, smith2013infectious, smith2014detecting} employed the Smith–Waterman algorithm, a dynamic programming technique, to compute alignment scores. Methods from bioinformatics for gene sequence alignment have also been adapted for intertextuality detection \citep{vesanto2017applying}.

\textbf{Entity matching}. In some types of literature, such as historical books, entities appear frequently and play an important role. Named entity information can be integrated to further optimize text similarity algorithms. \cite{deng-etal-2022-shu} employs a NER model to extract named entities from two books, and incorporates the overlap of entities into the text similarity calculation.

While overlap-based methods are highly interpretable, they struggle with non-literal linguistic relationships like synonymy and paraphrasing. Although they may seem outdated in the age of deep learning, these methods remain valuable in specific scenarios. Examples include low-resource languages, where training data for deep models is scarce, and damaged documents, where missing characters prevent the generation of meaningful vector representations.

Using advanced NLP techniques, discrete text can be transformed into dense vector representations, which help capture more complex and diverse intertextual relationships. This paper examines three types of vector similarity—word vector similarity, sentence vector similarity, and topic vector similarity—and discusses the application of generative large language models in intertextual detection.

\textbf{Word Vector Similarity.} This method employs language models to extract word vectors and then designs n-gram or sequence matching rules based on these vectors. Matching rules often need to be dynamically adjusted depending on the language and data type. For example, \cite{burns2021profiling} used word vectors extracted from language models to perform intertextual matching of bigrams in Latin literature. Specifically, for the four words across two bigrams, it computes the cosine similarity of their word vectors in a crosswise manner, then takes a secondary average based on the maximum similarity value and the average of the other three.

\textbf{Sentence Vector Similarity.} A sentence conveys a complete thought and is often used as a matching unit. Sentence vectors are typically extracted using deep neural networks. There are two primary methods for comparing sentences: one encodes each sentence separately to obtain its vector before calculating similarity \citep{cheng2023, duan2023disentangling}; the other feeds both sentences into a model that directly outputs a similarity score \citep{Liang2021guji, yu-etal-2022-ji}. The latter allows for richer interaction between sentences but demands labeled data for training and is more computationally intensive. Therefore, for large-scale corpora, the separate encoding approach is more suitable and practical, as it can readily leverage open-source sentence representation models.

\textbf{Topic Vector Similarity.} Modeling at the topic level aids in detecting broad, implicit intertextual connections. For instance, \cite{scheirer2014thesense} applied topic modeling to capture this type of overall intertextuality. TF-IDF serves as a classical method for textual similarity at the topic level. \cite{deng-etal-2022-shu}, for example, used n-grams as features to construct TF-IDF vectors.

Furthermore, some studies combine multiple approaches by designing hybrid metrics. These include: blending n-gram overlap, sequence matching, and semantic similarity to enhance metaphor recognition \citep{manjavacas2019feasibility}; integrating TF-IDF with entity matching \citep{deng-etal-2022-shu}; and deriving an intertextuality metric from a weighted sum of n-gram rules, character-level, and sentence-level semantic matching \citep{li-etal-2022-zhen}. 

\textbf{Generative Large Language Models}. Generative LLMs have achieved notable success across various text-related tasks, yet their use in intertextuality computation is still nascent. The potential of LLMs is considerable: Can prompt learning enable unsupervised, customized detection of intertextual links? How can their generative abilities be applied in innovative ways? For example, \cite{hatzel2024story} utilized prompted LLMs to derive text representations intended to capture narrative structure. \cite{umphrey2024investigating} proposed an approach where documents are input into an LLM, which then outputs detected intertextual pairs in natural language. This output is subsequently analyzed by domain experts. In this framework, the LLM serves as an intelligent assistant that generates a preliminary analysis, while experts apply their specialized knowledge to validate the findings. However, this method is not suitable for large-scale applications and is confined to case-specific studies. \cite{karjus2025machine} uses LLMs dialogically to determine whether a given text is related to a specific concept, which resembles concept-specific tracking rather than detecting intertextuality between two documents.

\subsubsection{Threshold}
The threshold is used to filter out intertextual pairs that satisfy similarity criteria. When annotated intertextual data is available, a supervised model can automatically determine an appropriate threshold. However, in most intertextuality studies, constructing an annotated dataset is costly. There are two main approaches for setting the threshold: empirical determination and manual evaluation-assisted determination.

The empirical method is simple and convenient: researchers design similarity scoring rules and set a threshold based on observation of the data. For example, \cite{cheng2023} retained intertextual pairs with a sentence vector cosine similarity greater than 0.9. \cite{duan2023disentangling} first recalled the top 100 candidate sentences for each sentence and pooled them together, then filtered out the top 1\% most similar sentence pairs. The drawback of this approach is its reliance on the researcher's subjective judgment. 
The manual evaluation-assisted method is more rigorous. For instance, \cite{wilkerson2015tracing} had human evaluators classify the intertextuality of retrieved pairs into six levels, and then trained a classifier to identify a suitable threshold.

Research objectives should also be taken into account. In general, if the goal is to review retrieved parallel intertexts case by case, then recall is more important, and a more lenient threshold should be chosen to yield more results. If the aim is to use the identified results for quantitative cultural analysis, then precision is prioritized, and a stricter threshold should be adopted to ensure more accurate retrieval.

\subsubsection{Evaluation Metric}

Parallel intertext detection is related to fields such as information retrieval, text matching, and text similarity; accordingly, its automatic evaluation metrics are also similar, including Recall, Precision, F1-score, and AUC/ROC curves.

In studies that focus on the effectiveness of intertext detection algorithms—especially when a benchmark exists with all intertexts in the dataset annotated, as in \cite{ganascia2014automatic, moritz-etal-2016-non, burns2021profiling, scheirer2014thesense, li-etal-2022-zhen}—standard methods can be used to compute these metrics. However, when the emphasis shifts to applying intertext algorithms to entirely new data, the calculation scheme for evaluation metrics must be designed. This section explains how to design such a scheme for recall and precision in the absence of gold-standard labels. The F1-score and AUC/ROC curves can then be derived from these two metrics.

Recall measures the proportion of known intertexts that the algorithm successfully retrieves. This metric is applicable whenever annotated intertext pairs are available; the annotations need not cover the entire dataset, as recall can be computed on the annotated subset. Such annotations may come from existing humanities research. For instance, \cite{forstall2014modeling} used a set of intertexts between two Latin epics (Virgil's \textit{Aeneid} and Lucan's \textit{Pharsalia}) compiled by scholars to evaluate recall, calculating how many of these known intertexts were retrieved by the algorithm.

Precision measures the proportion of the algorithm's detected intertexts that are correct. In large-scale experiments, which may yield millions of candidate pairs, it is common to sample a subset for manual evaluation. When comparing the precision of different algorithms, the total number of retrieved intertexts should be kept the same. This can be achieved by taking the top-k results by similarity score or by adjusting the detection threshold. For example, \cite{duan2023disentangling} used three retrieval methods to obtain the same number of intertext pairs and then compared their precision manually.

As the threshold for intertext retrieval changes, there is a trade-off between recall and precision, making it essential to control relevant variables to prevent misleading conclusions.
Additionally, intertextuality can manifest across multiple dimensions, such as semantics, syntax, prosody, and themes. During annotation, the specific type of intertextuality to focus on should be determined according to the research question.

\subsubsection{Acceleration}
Since intertextuality detection is an $O(n^2)$ problem, it faces a challenge of computational consumption. When dealing with large datasets, the computational cost becomes prohibitively high. To mitigate this issue, prior studies have introduced various strategies to reduce memory consumption and improve runtime efficiency. In n-gram matching, hashing is a widely adopted and effective technique. It converts text strings into hash values, thereby lowering memory requirements \citep{wilkerson2015tracing, linder2020text, smith2013infectious, smith2014detecting}. These hash values can then be filtered using specific rules—such as maximum and minimum pruning, TF-IDF weighting, and Winnowing—to exclude texts that are clearly not parallel intertexts.

For vector matching algorithms, integrating specialized vector retrieval acceleration tools has proven effective. A notable example is \textit{Faiss} \citep{johnson2019billion}, an industrial-grade tool designed for large-scale vector similarity search. It employs dimensionality reduction techniques to speed up the matching process while preserving high accuracy. With such tools, intertext detection can be performed at millions \citep{duan2023disentangling} or even tens of millions \citep{wang2024evol} level, using consumer-grade computing resources. Furthermore, methodologies from Retrieval-Augmented Generation (RAG), such as text chunking, embedding, and efficient large-scale retrieval, also offer valuable approaches for intertextuality detection.

\subsubsection{Meaningful Intertext}
Research on intertextuality often focuses on intertexts that convey substantive meaning, while disregarding those that are meaningless or overly commonplace. In this process, stop word lists are commonly employed, and word frequency serves as a key criterion for filtering out high-frequency textual elements. Several representative methods are summarized as follows:
\begin{itemize}
\item \textbf{Rule-based Filtering.} A straightforward strategy involves establishing explicit rules to exclude texts that do not meet predefined criteria. For instance, \cite{ganascia2014automatic} requires the number of high-frequency words in a sequence to exceed a set threshold. 
\item \textbf{Scoring Function.} A scoring function can be designed to evaluate the quality of intertexts. For example, in the case of parallel intertexts, methods may involve up-weighting rarer shared words \citep{forstall2014modeling}, down-weighting stop words \citep{li-etal-2022-zhen}, or combining both strategies \citep{sturgeon2018unsupervised}.
\item \textbf{Additional Discriminator.} Introducing newly annotated data and training a separate model to assist in judgment can be effective, though often costly. For example, \cite{burgess2016legislative} used annotated data to train a model for predicting whether a text is meaningful.
\end{itemize}
The underlying aim of these approaches is to disregard high-frequency, meaningless textual content.

\subsection{Normalization}
The count of detected parallel intertexts cannot be directly used to compare the strength of intertextuality across different text sets, because they are influenced by total text length: longer texts naturally yield higher counts, making comparisons invalid. Using unnormalized intertext counts to measure preferences or influence can lead to erroneous conclusions. Therefore, normalization is essential during intertextuality analysis to obtain comparable metrics. This section outlines three commonly used normalization techniques.

\textbf{Min-Max Normalization}.
This approach is often applied in one-to-many scenarios. It subtracts the minimum value first, then divides by the difference between the maximum and minimum values. For example, \cite{hartberg2017sacred} used Min–Max normalization on text reuse counts between text from one church's website to various parts of the Bible, thereby measuring differences in preference among books of the Bible. The normalization formula is: 
\begin{equation}
\frac{100(x-A)}{B-A}
\end{equation}
where $x$ is the count of text reuse and $A$ and $B$ are the minimum and maximum of the counts, respectively. 
\cite{cheng2023} used a similar approach to compare the preferences of poets Li Bai and Du Fu for different authors in the poetry anthology \textit{Wen Xuan}. 

\textbf{Unilateral Normalization}. This method is suitable for analyzing intertextuality within one set of texts, or for comparing multiple text sets against a single reference set. \cite{sturgeon2018unsupervised} used the ratio of characters involved in parallels to the total characters in a book to measure its intertextual strength:
\begin{equation}
\operatorname{par}(A, B)=\frac{\text { characters of } A \text { in } A:B \text { parallels }}{\text { characters in } A}
\end{equation}
\cite{buchler2013measure} applied a similar ratio at the sentence level to measure intertextual strength between ancient Greek philosophical and historical works (e.g., by Aristotle and Plato) and later texts.

\textbf{Bilateral Normalization}.
This method uses the length of two texts involved in intertextual computation to standardize the intertext quantity. This is suitable for all scenarios. \cite{sturgeon2018digital} incorporated the lengths of both texts into the denominator to derive a more general intertextuality metric:
\begin{equation}
\text { similarity }_{\text {ngram }}=\frac{\text { n-grams }_{\mathrm{AB}}}{\text { length }_{\mathrm{A}}+\text { length }_{\mathrm{B}}}
\end{equation}
This method uses character counts rather than intertext pairs, hence the additive denominator. Modeling results in the form of intertextual pairs can be adapted to other methods. 
Studies in both Chinese \citep{duan2023disentangling} and Latin \citep{gawley2017comparing} have used a normalization method based on the number of sentences in both texts. When measuring the intertextuality between two books, assuming the number of sentences in the two books is $n_i$ and $n_j$, and $x_{ij}$ intertextual pairs are detected between them, the normalized intertextuality $I_{ij}$ is:
\begin{equation}
I_{ij}=\frac{x_{ij}}{n_i * n_j}
\end{equation}
This scaling method is highly versatile: it can measure intertextuality between any two texts and supports cross-text comparisons. It is also mathematically sound: if intertexts are uniformly distributed, the number of detected pairs is proportional to the product of the text lengths.

The inherent characteristics of the data must be considered before normalization. For example, differences in punctuation conventions and linguistic expression across genres can affect sentence or character counts used in normalization; special handling may be needed when such differences significantly influence results. Additionally, when intertextual links are stored pairwise, duplicate texts can cause the number of detected pairs to grow exponentially. Thus, it is advisable to deduplicate texts beforehand. In summary, the selection and adjustment of normalization methods in intertextual analysis should be guided by the data properties, methodological assumptions, and the specific analytical task.

\subsection{Intertextual Feature Similarity}
Methods for detecting parallel intertextuality are more effective when analyzing textual content features and less suitable for capturing stylistic features. In such cases, machine learning-based anomaly detection algorithms provide a practical alternative. For example, \cite{dexter2017quantitative} trained one-class Support Vector Machine (SVM) anomaly detection models to distinguish between quoted material and original writing in the historical works of Livy. The original texts served as positive training examples. After training, recognized quoted passages were input into the model to measure their divergence. Another SVM model was trained to evaluate which texts resemble Livy's original writing, using his authentic works as positive instances and other Latin prose as negative instances. Similarly, \cite{forstall2011evidence} employed a similar approach by treating the works of the Roman poet Catullus as positive examples to assess his similarity to other poets, thereby measuring his influence on later generations.

\subsection{Intertextual Network}
Intertextual connections among a large number of texts naturally form a network structure, which is a common framework for intertextual analysis and visualization. Intertextual networks can be constructed along multiple dimensions. 

Sentence-level intertextual networks help identify clusters of similar texts and support the analysis of specific events, concepts, and topics. For example, \cite{cheng2023} built a sentence-level intertextual network of classical Chinese poetry and analyzed its subgraphs.

Book-level intertextual networks facilitate the study of cultural phenomena at the level of individuals or cultural groups. For instance, \cite{duan2023disentangling, sturgeon2018unsupervised} constructed book-level intertextual networks for classical Chinese texts, performing clustering and information diffusion analysis based on the network topology to explore questions related to scholars and schools of thought. \cite{romanello2016exploring} investigated composite intertextual networks by building and visualizing networks at the author, book, and passage levels.

When temporal information is incorporated, intertextual networks become directed graphs. For example, \cite{rockmore2018cultural} constructed a chronological directed graph using topic modeling of constitutions from various countries and time periods, then applied community detection algorithms to partition the graph.

Methods from social network analysis can be adapted for intertextual network analysis. \cite{cordell2015reprinting}, for instance, built a text reuse network of newspapers and computed metrics such as degree, betweenness centrality, and eigenvector centrality to quantitatively uncover dissemination patterns and influence within communities.

An intertextual network can be analyzed alongside external connections. For example, \cite{smith2013infectious} combined the news text reuse network with geographic information. By overlaying the text reuse network on a map with newspaper locations, the study enabled discussions about the paths, directions, and intensity of news dissemination. Similarly, \cite{linder2020text} integrated an intertextual network with a policy diffusion network, building an undirected network of U.S. states (nodes representing state-level bills) connected by text reuse links, alongside a directed network from prior research where edges indicated policy diffusion. By uncovering associations between the text reuse network and the actual policy diffusion network, the study validated the use of text reuse as a measure of policy similarity.

\section{Application}
Intertextuality modeling provides quantitative evidence to support research in the humanities and social sciences, helping to address abstract questions with data-driven insights. In what types of applications are intertextual connections utilized, and what kinds of research questions can they help answer? This article outlines five common scenarios: influence measurement, preference measurement, similarity measurement, writing habit analysis, and the analysis of cultural dissemination and evolution.

\subsection{Influence Measurement}
In literary creation, authors consciously or unconsciously draw on earlier texts. Intertextuality is frequently employed as an indicator of influence, helping assess the impact of individuals, works, schools, or social groups. When measuring the influence of one set of texts on another, the two sets generally need to follow a chronological sequence: the influencing texts are produced earlier, while the influenced texts are composed concurrently or later.

\cite{buchler2013measure} identified intertexts between 12 major ancient Greek philosophical and historical works and 7,200 other ancient Greek texts, using intertextuality to gauge the influence of these 12 classics. This study examined both the breadth and depth of influence, that is, the coverage and frequency of intertexts. For instance, Aristotle's \textit{On the Soul} showed the highest intertext coverage, indicating its content was widely adopted. Certain passages in Plato's \textit{Timaeus} were cited very frequently, reflecting the profound impact of its core ideas. The study also revealed how genre and time affect intertextual patterns: philosophical texts, being dense in argumentation, exhibited higher intertext coverage and stayed closer to original wording, whereas historical narratives had sparser intertext distributions, though hotspot events were often mentioned. And older texts generally had greater intertext coverage.

\cite{gawley2017comparing} calculated and compared the intertextuality of two key Roman authors, Cicero and Caesar, with texts from the Early and Late Empire. This allowed them to quantify the relative influence of these two authors across different time periods and genres. The results aligned with philological hypotheses: Late Empire authors cited Caesar more frequently, reflecting his stronger influence in historical epic and prose. Cicero was cited more often by Early Empire authors, especially in genres such as love poetry.

\cite{duan2023disentangling} use the strength of intertextuality to compare the influence of two philosophical school, Confucianism and Taoism. The findings suggested a clear Confucian influence in the political domain, while Confucianism and Taoism were more comparable in diverse literary works.

\cite{o2021text} used John Stuart Mill's borrowing and donation records from the London Library, then quantified text reuse between those sources and his published works, offering quantitative confirmation of the library's specific influence on the development of his ideas.

Beyond content-based intertextuality, stylistic similarity can also reflect influence. For example, \cite{dexter2017quantitative} used stylistic feature similarity to measure the influence of the historian Livy on later literary works. Similarly, \cite{forstall2011evidence} applied stylistic similarity to assess the influence of the Roman poet Catullus on Paul the Deacon's 8th-century Latin poem \textit{Angustae Vitae}.

The use of intertextuality to measure influence is also common in legal and political studies. \cite{burgess2016legislative} collected 550,000 bills and 200,000 resolutions from all 50 U.S. states, along with model bills from political lobbying organizations. They identified reused text between these bills and model bills. For different political groups, they calculated the rate at which their model bills were enacted into law in each state, visualizing the results on maps to reveal regional variations in political influence. 
\cite{wilkerson2015tracing, smith2014detecting} traced how policy ideas moved from proposals to formal laws, using the volume of text reuse to quantify the influence of Democratic and Republican parties on legislation.

\subsection{Preference Measurement}
Preference measurement can be understood as the inverse of influence measurement along a timeline. While influence measurement looks forward, preference measurement looks backward.

\cite{hartberg2017sacred} collected texts from the websites of six churches representing different denominations. By calculating the frequency of citations to the 66 books of the Bible in their sermons, they measured the citation preferences of conservative and progressive churches for various parts of the Bible.

\cite{duan2023disentangling} employed intertext distribution divergence to address controversial questions of authorship attribution. They computed the intertext distributions of both disputed and accepted sections of the \textit{Collected Works of Tao Yuanming}. The differences in how these sections are intertextual with Confucian and Taoist texts revealed distinctions in their underlying ideological preferences.

\subsection{Similarity Measurement}
Intertextuality can serve as an indicator of similarity in ideas, concepts, or writing styles. 
In studies of disputed authorship, analyzing intertextual features to measure the differences between authentic and spurious works can reveal distinctions in thought and composition. \cite{dexter2017quantitative} examined stylistic differences between eight tragedies written by Seneca and two disputed works. The study also applied anomaly detection to automatically separate original writing from quoted material. For instance, the Roman historian Livy extensively incorporated early historical sources through direct quotation, paraphrase, and vague allusion, making it challenging to systematically identify which passages were original and which were borrowed or adapted. A stylistic feature-based anomaly detection algorithm successfully distinguished Livy's original content from quoted material, confirming observable differences between the two.

The similarity between chapters within a book can help analyze its structure. \cite{sturgeon2018digital} visualized similarities among chapters of the \textit{Mozi} based on intertextual links. 

Additionally, \cite{linder2020text} employed text reuse to assess policy consistency. Using a dataset of U.S. state bills, the study identified text reuse across state legislation. By distinguishing bills introduced by Democrats and Republicans, it measured cross-state policy similarity within each party. Results showed that Republican legislators tended to promote a highly consistent conservative agenda across states, whereas policies proposed by Democratic legislators exhibited greater diversity.

\subsection{Writing Habits Analysis}
When authors quote or paraphrase earlier texts, they often modify the language and syntax. Examining the patterns of variation between detected intertexts can help interpret an author's rewriting tendencies. For example, the Chinese history book the \textit{Book of Han} directly incorporated material about the early Western Han from the \textit{Records of the Grand Historian}, while also making revisions and additions. \cite{deng-etal-2022-shu} quantitatively validated the traditional view that the \textit{Book of Han} tends to delete characters by statistically analyzing differences in character counts between intertextual pairs.

\cite{moritz-etal-2016-non} identified textual reuse of the Bible in Latin and Greek texts and studied the kinds of changes that occurred in these reused passages. Their research found that in Bernard of Clairvaux's Latin works, capitalizing the first letter of words at the start of a reused segment often corresponded to the original Biblical text, suggesting the author's tendency to use vocabulary from the Bible as a ``lead-in'' to introduce a reused segment.

Comparing the outcomes of different data processing and intertext detection methods applied to the same text can reveal variations in intertextual patterns. For instance, \cite{buchler2014towards} applied different preprocessing techniques (such as case normalization, similar character replacement, lemmatization, and synonym substitution) and text representation models (like unordered bag-of-words and ordered n-grams) to detect intertexts across different versions of the Bible. They observed that bag-of-words models are more effective at identifying semantically similar passages, while n-gram models perform better at detecting direct quotations. By comparing the recall rates of intertexts identified under different computational approaches across scriptural versions, the study evaluated the writing habits of various translators. For example, Young's Literal Translation showed very low n-gram similarity with other translations but relatively high bag-of-words similarity. This indicates that its translation adheres closely to Hebrew syntax rather than restructuring the text into fluent English, as other versions do. As a result, it yields different results under n-gram detection, yet still shares core vocabulary at the lexical level, allowing the bag-of-words model to capture similarities.

\subsection{Cultural Evolution Analysis}
Intertextuality provides a valuable framework for tracing the dissemination and transformation of ideas and cultures across time. The connection between texts can serve as microevidence to quantify the intensity of ideological dissemination.

\cite{rockmore2018cultural} employed topic links to connect a corpus of over 500 constitutions from diverse regions and historical periods. This approach facilitated an analysis of issues such as the periodization of constitutional development and the typical lifespan of constitutions.

\cite{duan2023disentangling} measured the rise and fall of philosophical influence over two millennia by analyzing the intertextuality between foundational texts from China's Hundred Schools of Thought (Axial Age) and later writings. The study also explored patterns of imitation and integration during the introduction of Buddhism into China.

\cite{smith2013infectious} integrated intertextual networks with spatiotemporal data to reveal how information diffusion is shaped by physical infrastructure (e.g., transport networks) and social dynamics. A key finding was the correlation between content type and dissemination speed: breaking news spread rapidly, whereas content with enduring value, such as poetry and philosophical works, exhibited a slower, more sustained spread pattern.

\section{Related Fields}
Intertextuality computation intersects with several related research areas. Text similarity represents a significant form of intertextuality, and citation is one of the concrete manifestations of intertextuality. While both have been extensively studied in their respective fields, certain distinctions remain between these concepts and the broader notion of intertextuality.

\textbf{Text Similarity}. 
Text similarity in natural language processing typically focuses on semantic similarity. Text Reuse and Text Alignment are two downstream tasks, and parallel intertextuality detection often adopts their methods. The difference is that, intertextuality includes any form of textual connection, such as semantics, style, theme, and structure. As a comprehensive concept, intertextuality allows for the incorporation of diverse computational techniques and analytical perspectives.

\textbf{Citation}. Citation analysis in the field of informetrics primarily focuses on explicit citation markers (direct citations), such as those in modern academic papers. Intertextuality research, however, concerns itself not only with direct citations but also, and more importantly, with unattributed borrowings (indirect citations) without bibliographic references. Research on citation recognition in ancient texts \citep{huang2021yinshu}, because it identifies specific cited works, can be said to identify citation relationships. If similar intertextual pairs are merely detected in parallel intertext detection, one cannot directly claim a citation relationship exists between them—they might be coincidentally similar without intentional citation.

\section{Platforms and Tools}
Online open-access projects have become a crucial medium for presenting intertextuality. While traditional intertextuality research culminated in printed documents for reading, digital intertextuality studies are more readily transformed into online platforms. 
By presenting detected intertextual links via online platforms, these systems offer readers a more accessible and convenient reading experience. 
These platforms integrate functions such as data retrieval, filtering, analysis, and visualization, allowing users to interact with the data. 
Meanwhile, packaging intertext detection algorithms into reusable computational tools enables other researchers to apply these methods directly, greatly facilitating the adoption of the intertextuality research paradigm among scholars. Platformization helps attract both academic and general-interest audiences, while tool-based digital intertextuality outputs create efficient bridges for interdisciplinary collaboration.

This section surveys browsing platforms and detection tools developed for Chinese, English, Latin, and Finnish intertextuality studies. The review focuses specifically on platforms and tools designed for humanities and social sciences research; general-purpose text similarity detection tools from the computer science field are beyond its scope.

\textbf{Chinese}. Digital platform for intertextuality is most extensive for Chinese, including intertext browsing and detection functions. 
The CTEXT platform\endnote{CTEXT Platform: \url{https://ctext.org/}} \citep{sturgeon2018unsupervised, sturgeon2018digital, sturgeon2019} not only provides online browsing of detected intertexts within its digital library but also offers an online interactive interface for text reuse detection\endnote{CTEXT Text Reuse Detection: \url{https://ctext.org/plugins/texttools/\#similarity}}. Users can paste their own text into the interface, and the platform returns all similar intertexts within the input text. It operates based on n-gram matching, allowing users to adjust hyperparameter settings. 
The Evol project\endnote{Evol Platform. V1: \url{http://evolution.pkudh.xyz/book}; V2: \url{https://ca.pkudh.net/}} \citep{duan2023disentangling,wang2024evol} provides online browsing and analysis for 14 million pre-detected intertextual pairs within the built-in corpus. It provides further analysis and visualization functions at levels like book, chapter, and sentence, but is limited to the built-in corpus. The project has open-sourced the training and detection code based on deep neural networks\endnote{Evol Code: \url{https://github.com/CissyDuan/Evol}}, enabling training and detection on users' own data. 
The Knowledge Graph platform\endnote{Knowledge Graph Platform: \url{https://cnkgraph.com/Tool/Referring}} provides an online interactive interface for poetry-oriented intertext detection. Users paste their text, and the platform returns similar texts from its built-in poetry database. 
Chinese Text Reuse\endnote{Chinese Text Reuse Code: \url{https://github.com/vierthaler/chinesetextreuse}} \citep{vierthaler2019blast} is an open-source detection tool based on the BLAST algorithm. It provides Python code requiring local deployment. 
Additionally, BERT-CCPoem\endnote{BERT-CCPoem Model: \url{https://github.com/THUNLP-AIPoet/BERT-CCPoem}} is an open-source poetry sentence embedding model. Based on this model, methods for detecting similar poetry lines can be easily developed \citep{cheng2023}.

\textbf{English}. Viral Texts\endnote{Viral Texts Platform: \url{https://viraltexts.org/}} \citep{smith2013infectious, smith2014detecting, smith2015computational} is a visualization platform for 19th-century American newspaper data. It visualizes pre-detected text reuse results using a graph structure and supports simple interactions on the graph. Clicking a node displays the names, links, and statistical information of other newspapers with intertextual links to it, but does not show the reused text content. The team also open-sourced a similar text detection toolkit, passim\endnote{passim: \url{https://github.com/dasmiq/passim}}, which can be installed for use on one's own data. TextPAIR\endnote{TextPAIR Code: \url{https://github.com/ARTFL-Project/text-pair}} \citep{TextPAIR202201009} is a similar text detection toolkit supporting multiple languages. It operates based on n-gram similarity matching and requires local installation. It has been successfully applied in humanities research involving both Chinese and English texts.

\textbf{Latin \& Greek}. The Tesserae platform\endnote{Tesserae Platform: \url{https://tesserae.caset.buffalo.edu/}, \url{https://tesseraev3.caset.buffalo.edu/}} \citep{coffee2012intertextuality, forstall2014modeling, forstall2011evidence, scheirer2014thesense, coffee2013tesserae} supports the identification and filtering of intertextual parallels across Latin and Greek corpora. It allows users to select a specific subset of the built-in corpus to examine intertextual connections within that selection. The platform also offers various filtering and sorting functions, enabling customized research workflows. Tesserae has facilitated many scholarly investigations: some researchers have used its detected intertexts as a basis for further analysis \citep{bernstein2015comparative, gawley2017comparing}, while others have treated its outputs as a baseline for evaluating new computational methods \citep{shang2021improving, burns2021profiling}.

\textbf{Finnish}. A Finnish platform\endnote{Finnish newspapers: \url{http://comhis.fi/clusters}} \citep{vesanto2017applying, salmi2020reuse, vesanto2017system} focuses on browsing and retrieving intertexts in Finnish newspapers from 1771–1910. It allows users to search for individual matches and clusters of texts, and can be saved in TSV format for continued analysis.
The Runoregi Platform\endnote{Runoregi Platform: \url{https://runoregi.rahtiapp.fi}} \citep{janicki2023exploring} can be used to browse intertexts in Finnish tetrametric poetry. It provides visualization tools allowing users to quickly read similar texts, finding corresponding lines and paragraphs.

As the number of intertextual links grows exponentially with the number of texts, developing and maintaining intertextuality platforms face challenges. For platforms displaying up to a million intertextual links, operational costs remain acceptable. Displaying larger-scale data not only incurs potentially prohibitive costs but also affects the response speed of calculations like filtering and analysis. Since intertext detection is an $O(n²)$ problem, providing free, full-scale intertext detection for large-scale texts is infeasible. Consequently, existing platforms either restrict the scale of detection or provide only toolkits. Despite these limitations, these platforms and tools have played a vital role in promoting this field.

\section{Challenge and Prospect}

Computational methods have markedly increased the efficiency of intertextual links detection, marking a substantial advance over traditional manual extraction. Nevertheless, limitations persist in methodology, data, and evaluation. To establish a rigorous framework for quantitative intertextuality research, these shortcomings must be addressed.

First, current techniques for detecting intertextuality remain imperfect. Some forms of intertextuality—such as direct semantic similarity—are readily observable. Others, including allusions and metaphors, are more abstract and elusive, posing continued challenges for automated identification. Furthermore, deep learning methods have broadened the potential of intertext detection, but their black-box nature often lacks interpretability.

Second, the evaluation criteria are often ambiguous. Intertextuality is an inherently fluid concept with diverse forms of textual relationships that may be prioritized differently depending on the research focus. For example, stylistic analysis may emphasize functional words, while studies on idea dissemination may prioritize meaningful content, requiring filtering out irrelevant functional words. Given the wide range of research purposes, establishing unified evaluation standards remains difficult.

Third, there is a shortage of reliable benchmark datasets. While small-scale, domain-specific datasets have been annotated for particular studies, the absence of clear and consistent evaluation standards complicates the creation of comprehensive, large-scale benchmarks. Moreover, because intertextuality manifests differently across research contexts, strong performance on one dataset does not guarantee effectiveness in others. \cite{fu2025ancienttrd} leveraged the generative power of LLMs to create positive and negative training samples for text representation models, thereby constructing a benchmark for intertextuality detection. \cite {karjus2025machine} used LLMs to generate test data for calculating the accuracy of intertextual detection. The collaboration between LLM generation and expert verification significantly enhanced the efficiency of benchmark development, presenting a promising paradigm.

Despite these challenges, the field remains promising. This paper contends that quantitative intertextuality research is poised for breakthroughs in the next five years, particularly in three areas: cross-lingual intertextuality, multimodal intertextuality, and narrative structure analysis.


\textbf{Cross-lingual Intertextuality}. Current research in digital intertextuality is largely confined to a single language. Investigations into cross-lingual intertextuality can connect different cultural systems by tracking the dissemination and transformation of ideas across languages. Initial investigations have explored cross-linguistic methods. For example, \cite{karjus2025machine} conducted preliminary tests on the ability of generative LLMs in cross-language intertextuality detection. But in-depth research is still rare. In the field of machine translation, cross-lingual sentence alignment methods are used to build parallel corpora, which can be transferred to cross-lingual intertextuality studies. For instance, \cite{liu2023bertalign} studied cross-lingual text alignment using a high-quality, manually aligned Chinese-English literary corpus.

\textbf{Multimodal Intertextuality}. The concept of intertextuality can be generalized to images and audio. For example, \cite{resler2021deep} employed Convolutional Neural Networks (CNNs) to extract features from images of artifacts to identify similar items, assisting archaeological research. The multimodal deep learning model provides the technical foundation for cross-modal intertextuality studies, which is still a largely open field.

\textbf{Narrative Structure}. Methods for quantifying narrative similarity remain limited. Some studies have explored story embeddings and established benchmarks. \cite{chaturvedi2018have} constructed a dataset of 577 movie summaries, assessing narrative similarity from plot-related angles (using bag-of-words models for events and entities) and character perspectives (aligning attributes such as gender and role relationships). \cite{hatzel2024story, hatzel2024tell} assembled a larger dataset comprising 96,831 summaries representing 29,505 stories. However, these benchmarks rely on summary texts rather than complete stories, and the datasets mainly include different versions of the same story (e.g., translations or remakes), as opposed to identifying shared narrative structures across different stories (e.g., \textit{Hamlet} and \textit{The Lion King}). The methods also tend to focus heavily on semantic similarity, with inadequate modeling of narrative architecture. Modeling narrative intertextuality continues to be an unresolved challenge.

\section{Conclusion}

This paper presents a review of quantitative intertextuality research, synthesizing existing studies from three key perspectives: research data, computational methods, and applications. 

In terms of data, this paper highlights two scenarios particularly suitable for intertextuality analysis: ancient texts, which often lack consistent writing conventions and background information; and texts with similar content that exist in multiple versions, which can be used to investigate variations between different versions. Intertextuality analysis applied to such materials has proven to be an effective research methodology.

Regarding methods, this paper provides a detailed overview of the technical pipeline for intertextuality modeling. This includes data preprocessing, the selection of appropriate intertextual features, the detection of parallel intertextual pairs, and several advanced modeling techniques such as normalization, intertextual feature similarity measurement, and intertextual network modeling. These approaches quantify intertextuality from various perspectives and reflect an evolution from classical statistical methods to deep learning.

For applications, this paper summarizes the types of research questions that can be investigated through quantitative intertextuality modeling. These include using intertextual strength to measure influence, preference, and similarity; analyzing writing habits through variations in intertextuality; and employing intertextual links as clues to trace cultural dissemination and evolution. Additionally, this paper investigates available online platforms and computational tools that support intertextual browsing and detection.

Finally, this paper outlines current limitations of quantitative intertextuality studies, such as technological constraints, ambiguous evaluation standards, and the lack of benchmark datasets, along with promising future research directions, including cross-lingual studies, multimodal intertextuality, and narrative structure similarity analysis.

This paper provides a roadmap for quantitative intertextuality research. Driven by advances in computer technology, quantitative intertextuality studies hold significant potential for methodological and applicational development. More precise, diverse, and large-scale intertext detection methods can be anticipated, and intertextuality modeling approaches are expected to be transferable to various types of humanities and social sciences research.



\section{Competing Interests}
No competing interest is declared.

\section{Author Contributions Statement}
Siyu Duan is the sole contributor to this work.
\section{Data Availability Statement}
No new data were generated or analysed in support of this research.

\clearpage
\theendnotes
\clearpage

\bibliographystyle{abbrvnat}
\bibliography{reference}


\end{document}